\documentclass[11pt]{article}

\usepackage{acl}

\usepackage[utf8]{inputenc}
\usepackage[T1]{fontenc}

\usepackage{amsmath}
\usepackage{amssymb}

\usepackage{booktabs}
\usepackage{multirow}
\usepackage{array}
\usepackage{tabularx}

\usepackage{graphicx}
\usepackage{float}
\usepackage{subcaption}  

\usepackage[dvipsnames]{xcolor}
\usepackage{colortbl}
\definecolor{rowgray}{gray}{0.93}

\usepackage{cleveref}    

\usepackage{enumitem}
\setlist{noitemsep, topsep=3pt}

\title{Quantization Effects on Bangla Language Understanding\\
       in Large Language Models: A Systematic Evaluation}

\author{
  \textbf{Ismail Hossain}\thanks{Equal contribution.} \quad
  \textbf{Nafi Ullah Shafin}$^{*}$ \quad
  \textbf{Mohammad Abdullah Al Mumin} \\[2pt]
  Institute of Information and Communication Technology (IICT) \\
  Shahjalal University of Science and Technology (SUST), Sylhet, Bangladesh \\[2pt]
  \texttt{\{ismail33, nafi04\}@student.sust.edu, mumin-cse@sust.edu}
}
\begin{document}
\maketitle

\begin{abstract}
Post-training quantization lowers the memory footprint of Large
Language Models (LLMs) and speeds up inference, which is why it is
now common for on-device deployment. Most of what we know about its
effects, however, comes from English benchmarks. It is not clear
whether the same holds for morphologically complex, low-resource
languages such as \textbf{Bangla}, and this gap is what we address
here. We evaluate three model families---\textbf{Qwen-2.5-7B},
\textbf{LLaMA-3.1-8B}, and \textbf{GPT-OSS-20B}---in full precision
and in three quantized formats (GPTQ-Int8, GPTQ-Q8, GGUF-W8A16)
across five Bangla natural language understanding benchmarks (Bangla
MMLU, CommonsenseQA-BN, OpenBookQA-BN, PIQA-BN, and BoolQ-BN), using
zero-shot evaluation through \texttt{lm-evaluation-harness}. To our
knowledge this is the first controlled comparison of quantization
formats on Bangla NLU. The three families do not respond the same
way: GPT-OSS loses up to 57.35\% accuracy on reasoning-heavy tasks
under GGUF-W8A16, while Qwen and LLaMA hold steady under GPTQ, and in
a few cases the quantized version edges out the full-precision one.
BoolQ-BN, a comprehension task, stays stable across all three
families regardless of format. Taken together, these results suggest
quantization can work well for Bangla deployment, but the choice of
architecture and quantization method matters more than the bit width
alone. We discuss what this means for practitioners choosing a model
to run on constrained hardware.
\end{abstract}

\section{Introduction}
\label{sec:intro}

Large language models such as GPT-4, LLaMA, and Qwen now deliver
strong performance across a wide range of NLP
tasks~\citep{jin2024comprehensive, li2024evaluating}, but running
them is expensive. Frontier models need tens to hundreds of
gigabytes of GPU memory, which puts them out of reach in places where
computing infrastructure is limited, including much of South Asia,
home to more than 230 million Bangla speakers.

\emph{Post-training quantization} (PTQ) is one way around this
problem~\citep{lang2024quantstudy, lee2025lrq}. Dropping weight
precision from 16-bit floating point to 8-bit integers or lower cuts
memory use by 2$\times$--8$\times$ and speeds up inference by
2$\times$--5$\times$, with little extra engineering effort, enough to
make on-device deployment realistic on laptops, phones, and edge
servers. Methods such as GPTQ and AWQ push this further, using
second-order weight information or activation-aware compression to
limit accuracy loss~\citep{frantar2022gptq, liu2025comprehensivequant}.

Nearly all of this evaluation work, though, happens on English
benchmarks: MMLU, GSM8K, HumanEval~\citep{ding2024evaluating,
li2024evaluating}. That leaves a question we have not seen answered
in the literature: does quantization error compound differently for
a linguistically complex, low-resource language like Bangla? Bangla's
agglutinative morphology, its conjunct consonant clusters, and the
grapheme composition rules of its Unicode script all complicate
tokenization in ways that could plausibly make representation
degradation worse under compression~\citep{bhowmik2025evaluating,
nahin2025titullms}.

\paragraph{Research questions.}
\begin{enumerate}[label=\textbf{RQ\arabic*.}, leftmargin=*]
  \item How does post-training quantization affect LLM accuracy on
        Bangla NLU benchmarks?
  \item Which quantization formats (GPTQ-Int8, GPTQ-Q8, GGUF-W8A16)
        best preserve Bangla accuracy?
  \item Are specific Bangla task categories (reasoning, commonsense,
        comprehension) differentially sensitive to quantization?
\end{enumerate}

\paragraph{Contributions.} This paper makes four contributions.
\begin{enumerate}[leftmargin=*]
  \item We run the first controlled, multi-family comparison of PTQ
        effects on Bangla NLU that we are aware of, covering 15
        model--benchmark pairs across five task categories.
  \item We find a sharp split between model families: GGUF-quantized
        GPT-OSS degrades by up to 57.35\% on reasoning tasks, while
        GPTQ-quantized Qwen and LLaMA show under 1.5\% degradation,
        and sometimes a small improvement, across all five benchmarks.
  \item Reasoning and commonsense benchmarks turn out to be far more
        fragile under quantization than reading comprehension, in the
        Bangla setting specifically.
  \item We turn these results into concrete recommendations for
        choosing a model architecture and quantization format for
        on-device Bangla NLP.
\end{enumerate}


\section{Related Work}
\label{sec:related}

\subsection{Multilingual LLMs and Low-Resource Languages}

GPT, LLaMA, Qwen, Mixtral, and Mistral are all pretrained on large
multilingual corpora and generalize well in zero-shot settings for
the languages that dominate their training data. Performance drops
off for low-resource languages, and the usual reasons are sparse
token coverage, tokenization mismatches, and morphological
complexity~\citep{bhowmik2025evaluating}. Bangla runs into all three:
agglutinative morphology, conjunct-consonant orthography, and a
comparatively small web-scale footprint leave even strong
multilingual models underperforming on Bangla
tasks~\citep{bhowmik2025evaluating, nahin2025titullms}.

Two efforts try to close this gap directly. BanglaBERT is a
BERT-style encoder trained on curated Bangla text, and IndicLLMs
extends multilingual pretraining to South Asian scripts more broadly.
The largest recent effort is TituLLMs~\citep{nahin2025titullms}, a
family of Bangla-native decoder-only models built with Bangla-aware
subword tokenizers and benchmarked across CommonsenseQA-BN, MMLU-BN,
PIQA-BN, and BoolQ-BN. The TituLLMs authors argue that tokenizer
design and corpus quality are what mainly control Bangla NLU
performance, which raises a question they do not answer: if
tokenizer and corpus quality matter this much, does compressing the
model after training erode that advantage? We did not find any prior
work that looks at quantization and Bangla-specific linguistic
properties together.

\subsection{Post-Training Quantization for LLMs}

Post-training quantization (PTQ) lowers model weight precision after
training is already done, which sidesteps the cost of
quantization-aware retraining. The simplest version is uniform INT8
quantization, applying per-tensor or per-channel linear scaling.
GPTQ~\citep{lee2025lrq} goes further, minimizing layer-wise
reconstruction error using approximate second-order information, and
AWQ (Activation-Aware Weight Quantization) refines this again by
preserving the weights that activation magnitudes mark as
salient~\citep{liu2025comprehensivequant}. GGUF, the serialization
format \texttt{llama.cpp} uses for CPU/GPU inference, supports
several internal bit depths (Q4\_K\_M, Q5\_K\_S, Q8\_0, W8A16) and is
common in on-device deployment pipelines.

Across these methods, quantization typically cuts memory use by
2$\times$–8$\times$ and speeds up inference by 2$\times$–5$\times$,
which is what makes consumer laptops and edge hardware viable targets
for LLM deployment~\citep{lang2024quantstudy, liu2025comprehensivequant}.

\subsection{Evaluation of Quantized LLMs}

A handful of studies measure quantized LLM accuracy on English
benchmarks. \citet{li2024evaluating} evaluate multiple quantization
methods across MMLU, HumanEval, and GSM8K and find that INT4 causes
real degradation on complex reasoning tasks while INT8 stays mostly
safe. \citet{ding2024evaluating} build a generalization-aware
evaluation toolbox that stress-tests quantized models on
out-of-distribution examples. \citet{jin2024comprehensive} compare
GPTQ, AWQ, and SmoothQuant and find that higher-level cognitive tasks
(math, code) degrade more than factual retrieval under the same bit
reduction.

Every one of these studies works from English-only benchmarks, and
none of them touch a low-resource or morphologically complex
language. That is the gap this paper fills: we evaluate quantized
LLMs on five Bangla benchmarks spanning reasoning, commonsense
inference, and reading comprehension, which as far as we can tell has
not been done before.


\section{Methodology}
\label{sec:method}

We compare full-precision and quantized variants of three LLM
families across five benchmark datasets to see how post-training
quantization affects Bangla NLU. Within each family, quantization is
the only thing that changes between the two compared models; prompt
format, decoding strategy, and evaluation harness version are all
held fixed.

\subsection{Benchmark Datasets}
\label{sec:datasets}

We use five publicly available Bangla NLU benchmarks, all sourced
from the \texttt{hishab} organization on Hugging Face, spanning three
task categories: reasoning, commonsense inference, and reading
comprehension.

\paragraph{Bangla MMLU} (\texttt{hishab/bangla-mmlu}).
A translation of the English MMLU benchmark~\citep{hendrycks2021mmlu}
covering 57 subjects across science, humanities, and social sciences.
Tests broad encyclopaedic reasoning in a four-choice multiple-choice
format.

\paragraph{CommonsenseQA-BN} (\texttt{hishab/commonsenseqa-bn}).
A five-choice multiple-choice benchmark requiring implicit real-world
reasoning. Translated from the English
CommonsenseQA~\citep{talmor2019commonsenseqa} dataset.

\paragraph{OpenBookQA-BN} (\texttt{hishab/openbookqa-bn}).
Four-choice multiple-choice questions~\citep{mihaylov2018openbookqa}
requiring multi-step elementary science reasoning combined with
external factual knowledge.

\paragraph{PIQA-BN} (\texttt{hishab/piqa-bn}).
A two-option benchmark~\citep{bisk2020piqa} evaluating physical and
procedural commonsense reasoning (e.g., determining the correct
method to accomplish a practical task).

\paragraph{BoolQ-BN} (\texttt{hishab/boolq\_bn}).
A binary (Yes/No) reading comprehension benchmark~\citep{clark2019boolq}
where each question is accompanied by a short passage; the model must
extract and verify a factual claim.

Together, these five datasets provide a balanced coverage of task
difficulty and linguistic phenomena, consistent with the evaluation
suite used by TituLLMs~\citep{nahin2025titullms}. Per-benchmark
example counts are reported in \Cref{tab:datastats}
(\Cref{app:stats}).

\subsection{Models and Quantization Variants}
\label{sec:models}

We evaluate one full-precision baseline and one quantized variant per
model family, controlling for model size within each family.
\Cref{tab:models} summarises all evaluated checkpoints.

\begin{table}[H]
\centering
\caption{Evaluated models and quantization configurations.}
\label{tab:models}
\renewcommand{\arraystretch}{1.25}
\resizebox{\columnwidth}{!}{%
\begin{tabular}{llll}
\toprule
\textbf{Family} & \textbf{Variant} & \textbf{Format} & \textbf{Params} \\
\midrule
\multirow{2}{*}{Qwen}
  & Qwen2.5-7B-Instruct            & FP16     & 7B \\
  & Qwen2.5-7B-Instruct-GPTQ-Int8  & GPTQ-Int8 & 7B \\
\midrule
\multirow{2}{*}{LLaMA}
  & Meta-Llama-3.1-8B-Instruct     & FP16/BF16 & 8B \\
  & Meta-Llama-3.1-8B-GPTQ-Q\_8   & GPTQ-Q8   & 8B \\
\midrule
\multirow{2}{*}{GPT-OSS}
  & openai/gpt-oss-20b             & FP16      & 20B \\
  & gpt-oss-20b-ShiningValiant3-W8A16 & GGUF-W8A16 & 20B \\
\bottomrule
\end{tabular}%
}
\end{table}

\paragraph{Qwen family.}
Qwen2.5-7B-Instruct~\citep{qwen2024qwen25} is a multilingual
instruction-tuned model that already performs reasonably on Bangla.
We compare it against a GPTQ variant~\citep{lee2025lrq} that
compresses weights to 8-bit integers.

\paragraph{LLaMA family.}
Meta-Llama-3.1-8B-Instruct~\citep{dubey2024llama3} is compared
against a GPTQ-Q8 quantized variant. Its BF16 training and
grouped-query attention give us a useful second architecture to check
whether robustness patterns hold across model designs, not just
within one.

\paragraph{GPT-OSS family.}
GPT-OSS-20B~\citep{openai2025gptoss} is a large open-weight instruct
model. We compare it against a GGUF-W8A16 variant, serialized in the
GGUF format and loaded through \texttt{llama.cpp}, which is the kind
of deployment-optimized checkpoint most consumer inference engines
actually use.

\subsection{Evaluation Pipeline}
\label{sec:pipeline}

All evaluations run on EleutherAI's
\texttt{lm-evaluation-harness}~\citep{gao2021harness}, which is the
toolkit OpenAI, Meta, and most of the academic community already use
for reproducible LLM benchmarking. For each model, we load the
checkpoint (full-precision through
\texttt{transformers}~\citep{wolf2020transformers}, quantized through
\texttt{auto-gptq} or \texttt{llama.cpp}), query each benchmark in
\textbf{zero-shot} mode with no task-specific prompting or few-shot
demonstrations, select answers by comparing log-likelihoods over the
candidate choices (or over the ``Yes''/``No'' tokens for BoolQ-BN),
and record accuracy per benchmark. We keep structured JSON logs of
every run for post-hoc analysis; these will be released alongside the
code.

\paragraph{Reproducibility.}
Random seeds are fixed and evaluation uses greedy log-likelihood
selection, so there is no sampling variance to speak of. We did see
minor variance ($<$0.3\%) across re-runs on the same hardware for a
few benchmarks, small enough that it would not change any conclusion
we draw. Experiments ran on cloud GPU instances (NVIDIA A100/V100)
since our local hardware could not handle the larger checkpoints;
full environment specifications and total compute budget are
reported in \Cref{app:stats}.

\subsection{Evaluation Metrics}
\label{sec:metrics}

\paragraph{Accuracy.}
All five benchmarks use categorical prediction. Primary metric:
\begin{equation}
  \text{Accuracy} = \frac{\text{\# correct predictions}}
                         {\text{\# total questions}}
  \label{eq:acc}
\end{equation}

\paragraph{Performance Degradation.}
To quantify quantization impact we report both absolute and
relative degradation:
\begin{align}
  \Delta &= \text{Acc}_{\text{full}} - \text{Acc}_{\text{quant}}
  \label{eq:delta} \\[4pt]
  \Delta\% &= \frac{\text{Acc}_{\text{full}} - \text{Acc}_{\text{quant}}}
                   {\text{Acc}_{\text{full}}} \times 100
  \label{eq:delta_pct}
\end{align}
Negative values of $\Delta$ (i.e., the quantized model outperforms
the full-precision baseline) are reported without modification; we
discuss their likely interpretation in \Cref{sec:discussion}.


\section{Results}
\label{sec:results}

\begin{figure}[H]
  \centering
  \includegraphics[width=\columnwidth]{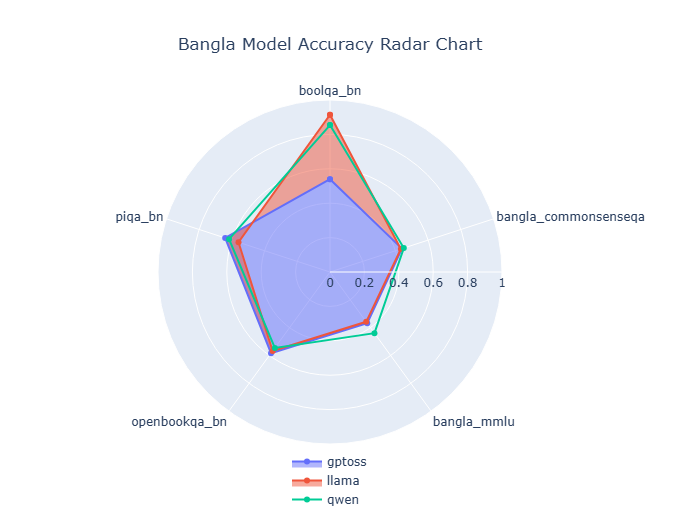}
  \caption{Radar charts of \emph{full-precision} model accuracy
           across five Bangla benchmarks. Each axis represents one
           benchmark; a larger shaded area indicates higher overall
           performance.}
  \label{fig:radar_full}
\end{figure}

\begin{figure}[H]
  \centering
  \includegraphics[width=\columnwidth]{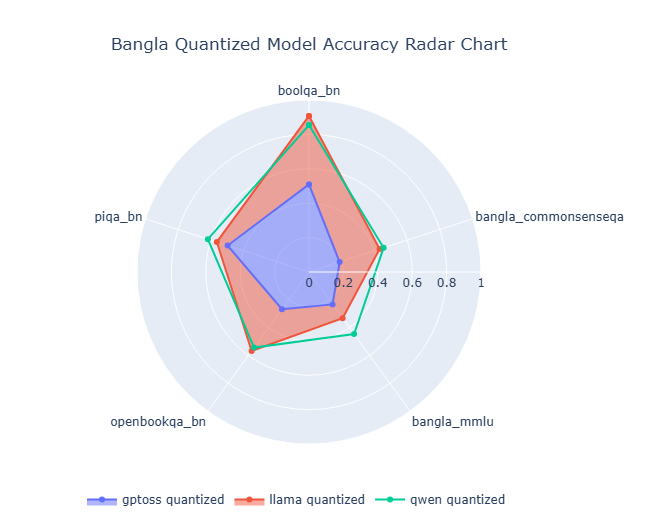}
  \caption{Radar charts of \emph{quantized} model accuracy on the
           same five benchmarks. Note the severe contraction of the
           GPT-OSS (blue) area on CommonsenseQA-BN and OpenBookQA-BN,
           contrasting with near-unchanged Qwen (green) and LLaMA
           (red) profiles.}
  \label{fig:radar_quant}
\end{figure}

We report zero-shot accuracy for all six model checkpoints (3
families $\times$ 2 precision variants) across all five Bangla
benchmarks. \Cref{tab:full_prec,tab:quant,tab:degradation} give the
raw accuracy scores and the degradation metrics computed from them.
Figures~\ref{fig:radar_full} and~\ref{fig:radar_quant} show the same
numbers as radar charts, which makes the multi-benchmark profile of
each model easier to compare at a glance.

\subsection{Full-Precision Baseline Performance}
\label{sec:results_baseline}

\begin{table}[H]
\centering
\caption{Zero-shot accuracy of full-precision models on five Bangla
         benchmarks. MMLU = Bangla MMLU; CSQA = CommonsenseQA-BN;
         OBQ = OpenBookQA-BN; PIQA = PIQA-BN; BoolQ = BoolQ-BN.}
\label{tab:full_prec}
\renewcommand{\arraystretch}{1.3}
\resizebox{\columnwidth}{!}{%
\begin{tabular}{lccccc}
\toprule
\textbf{Model} & \textbf{MMLU} & \textbf{CSQA} & \textbf{OBQ}
               & \textbf{PIQA} & \textbf{BoolQ} \\
\midrule
\rowcolor{lightgray}
GPT-OSS-20B  & 0.369 & 0.440 & 0.584 & 0.641 & 0.539 \\
LLaMA-3.1-8B & 0.358 & 0.433 & 0.567 & 0.559 & 0.914 \\
\rowcolor{lightgray}
Qwen-2.5-7B  & 0.440 & 0.451 & 0.545 & 0.617 & 0.854 \\
\bottomrule
\end{tabular}%
}
\end{table}

Qwen-2.5-7B has the best MMLU accuracy among the full-precision
models (0.440). LLaMA-3.1-8B, despite being the smallest model we
test, scores 0.914 on BoolQ-BN, which points to strong reading
comprehension independent of parameter count. GPT-OSS-20B leads on
PIQA-BN (0.641) and OpenBookQA-BN (0.584) but falls behind on
BoolQ-BN. None of the three families is dominant across the board,
which tells us each has real Bangla NLU capacity going into the
quantization comparison, not just on the benchmarks it happens to
lead.

\subsection{Quantized Model Performance}
\label{sec:results_quant}

\begin{table}[H]
\centering
\caption{Zero-shot accuracy of quantized model variants on the
         same five Bangla benchmarks. Format labels are given
         in \Cref{tab:models}.}
\label{tab:quant}
\renewcommand{\arraystretch}{1.3}
\resizebox{\columnwidth}{!}{%
\begin{tabular}{lccccc}
\toprule
\textbf{Model (Quant.)} & \textbf{MMLU} & \textbf{CSQA} & \textbf{OBQ}
                        & \textbf{PIQA} & \textbf{BoolQ} \\
\midrule
\rowcolor{lightgray}
GPT-OSS (GGUF-W8A16)   & 0.233 & 0.188 & 0.268 & 0.497 & 0.509 \\
LLaMA (GPTQ-Q8)        & 0.333 & 0.432 & 0.567 & 0.564 & 0.907 \\
\rowcolor{lightgray}
Qwen (GPTQ-Int8)       & 0.446 & 0.456 & 0.543 & 0.618 & 0.854 \\
\bottomrule
\end{tabular}%
}
\end{table}

The gap between families shows up immediately. GPT-OSS loses accuracy
across every benchmark except BoolQ-BN, where the drop is a
comparatively mild 5.6\%. LLaMA and Qwen stay close to their
full-precision scores throughout. Qwen's quantized variant actually
beats its own full-precision version on MMLU (+0.6\%), CSQA (+0.5\%),
and PIQA (+0.1\%); we come back to what this small reversal likely
means in \Cref{sec:discussion}.

\subsection{Performance Degradation Analysis}
\label{sec:results_degradation}

\Cref{tab:degradation} presents per-family, per-benchmark degradation
computed via \Cref{eq:delta,eq:delta_pct}.

\begin{table}[H]
\centering
\caption{Performance degradation ($\Delta$ and $\Delta\%$) due to
         quantization. Negative $\Delta$ indicates the quantized
         model \emph{outperforms} its full-precision counterpart.
         Bold values mark the largest absolute degradations per family.}
\label{tab:degradation}
\renewcommand{\arraystretch}{1.3}
\resizebox{\columnwidth}{!}{%
\begin{tabular}{llccccc}
\toprule
\textbf{Family} & \textbf{Benchmark}
  & \textbf{Full} & \textbf{Quant.}
  & \textbf{$\Delta$} & \textbf{$\Delta\%$} \\
\midrule
\rowcolor{lightgray}
\multirow{3}{*}{GPT-OSS}
  & Bangla MMLU      & 0.369 & 0.233 & \textbf{0.136} & \textbf{36.7} \\
  & CommonsenseQA-BN & 0.440 & 0.188 & \textbf{0.252} & \textbf{57.4} \\
\rowcolor{lightgray}
  & OpenBookQA-BN    & 0.584 & 0.268 & \textbf{0.316} & \textbf{54.1} \\
  & PIQA-BN          & 0.641 & 0.497 & 0.144          & 22.4          \\
\rowcolor{lightgray}
  & BoolQ-BN         & 0.539 & 0.509 & 0.030          & 5.6           \\
\midrule
\multirow{5}{*}{LLaMA}
  & Bangla MMLU      & 0.358 & 0.333 & 0.025          & 7.0           \\
\rowcolor{lightgray}
  & CommonsenseQA-BN & 0.433 & 0.432 & 0.002          & 0.4           \\
  & OpenBookQA-BN    & 0.567 & 0.567 & 0.000          & 0.0           \\
\rowcolor{lightgray}
  & PIQA-BN          & 0.559 & 0.564 & $-$0.004       & $-$0.8        \\
  & BoolQ-BN         & 0.914 & 0.907 & 0.007          & 0.8           \\
\midrule
\rowcolor{lightgray}
\multirow{7}{*}{Qwen}
  & Bangla MMLU      & 0.440 & 0.446 & $-$0.006       & $-$1.4        \\
  & CommonsenseQA-BN & 0.451 & 0.456 & $-$0.005       & $-$1.1        \\
\rowcolor{lightgray}
  & OpenBookQA-BN    & 0.545 & 0.543 & 0.002          & 0.4           \\
  & PIQA-BN          & 0.617 & 0.618 & $-$0.001       & $-$0.2        \\
\rowcolor{lightgray}
  & BoolQ-BN         & 0.854 & 0.854 & 0.000          & 0.0           \\
\bottomrule
\end{tabular}%
}
\end{table}

\paragraph{Key observations.}
\begin{itemize}
  \item \textbf{GPT-OSS falls apart on reasoning tasks.}
        CommonsenseQA-BN ($\Delta\% = 57.4$) and OpenBookQA-BN
        ($\Delta\% = 54.1$) lose more than half their accuracy.
        PIQA-BN drops 22.4\%. BoolQ-BN, at 5.6\%, is the only
        benchmark where GPT-OSS still performs reasonably after
        quantization.

  \item \textbf{LLaMA barely moves.}
        GPTQ-Q8 keeps absolute degradation under 1\% on every
        benchmark. OpenBookQA-BN shows no change at all, and
        PIQA-BN even ticks up by 0.4\%.

  \item \textbf{Qwen holds up best of the three.}
        GPTQ-Int8 does not cost Qwen any measurable accuracy on
        any benchmark, and three of the five (MMLU, CSQA, PIQA)
        come in slightly higher than full precision ($<$1.5\%).

  \item \textbf{BoolQ-BN is the one benchmark all three families
        handle well.}
        Every family keeps a competitive BoolQ-BN score after
        quantization: GPT-OSS drops the most (0.030), LLaMA and
        Qwen by 0.007 or less.
\end{itemize}


\section{Discussion}
\label{sec:discussion}

\subsection{Architectural Resilience to Quantization}

The clearest result in this study is that quantization robustness
splits by family and has little to do with task category.
LLaMA-3.1-8B and Qwen-2.5-7B keep nearly all of their accuracy under
8-bit GPTQ compression, while GPT-OSS-20B, the largest model in the
comparison, loses up to 57.4\% on the same benchmarks. If anything,
being bigger did not help here.

A few architectural and training factors could explain this, though
we cannot fully separate them with the data we have. GPTQ calibration
quality depends on the calibration set: the LLaMA and Qwen GPTQ
checkpoints were calibrated on diverse multilingual corpora, while it
is not clear the GPT-OSS GGUF checkpoint (W8A16) was calibrated on
anything beyond English. Qwen's dense, MoE-free architecture may also
spread representational load more evenly across layers, which would
reduce the per-layer sensitivity that GPTQ's Hessian-based correction
is designed to target~\citep{lee2025lrq}. . We flag this as a real confound rather
than a settled explanation, since GPT-OSS is also the only family in
our study without a GPTQ counterpart to compare against directly (see
Limitations).

\subsection{Task Sensitivity: Reasoning vs.\ Comprehension}

In all three families, the reasoning-heavy benchmarks
(CommonsenseQA-BN, OpenBookQA-BN, Bangla MMLU) degrade more than
reading comprehension (BoolQ-BN). English-benchmark work has found
the same pattern~\citep{jin2024comprehensive, li2024evaluating}; our
results are, to our knowledge, the first to show it holds for a
morphologically complex, low-resource language too.

One explanation is task structure. BoolQ-BN asks the model to check a
factual claim against a passage sitting right there in the prompt, a
fairly local, surface-level matching operation, and small weight
perturbations do not seem to disturb that signal much.
CommonsenseQA-BN and OpenBookQA-BN ask for something harder: pulling
in implicit world knowledge and chaining several reasoning steps,
which leans on long-range attention patterns and softmax
distributions that are more sensitive to the kind of small
perturbations quantization introduces~\citep{liu2025comprehensivequant,
lang2024quantstudy}.

\subsection{Linguistic Interaction and the Bangla Factor}

Going in, we suspected Bangla's linguistic properties might amplify
quantization-induced degradation relative to English. There is some
indirect support for that: GPT-OSS loses 57.4\% on CommonsenseQA-BN,
well above the 10–20\% range typically reported for GGUF quantization
on English MMLU~\citep{ding2024evaluating}. Bangla's agglutinative
morphology and Unicode grapheme clusters produce longer, less frequent
token sequences than English text does for the same content, which
spreads meaning across more token positions and gives precision
errors more room to compound across attention layers.

That said, this reading is confounded by the same uncertainty around
GPT-OSS's GGUF calibration data noted above. Isolating a genuine
Bangla-specific effect would need a controlled ablation that holds
the quantization method fixed and varies only the language. We treat
the language-amplification idea as a hypothesis this study points
toward, not one it confirms.

\paragraph{On negative degradation values.}
Qwen-Int8 comes out ahead of Qwen-FP16 on MMLU ($+$0.6\%), CSQA
($+$0.5\%), and PIQA ($+$0.1\%). One known cause of this in GPTQ
calibration is that the Hessian-guided weight reconstruction can
occasionally smooth out overfitting or numerical instability already
present in the FP16 weights, effectively acting as a mild
regularizer~\citep{lee2025lrq}. These margins are small enough
($<$1.5\%) to sit within normal measurement noise, so we read them as
a sign of strong quantization robustness rather than as a genuine
accuracy gain.

\subsection{Practical Implications for Bangla NLP Deployment}

These results translate into three concrete recommendations:

\begin{itemize}
  \item \textbf{Use GPTQ-Int8 or GPTQ-Q8 for on-device deployment.}
        Qwen-2.5-7B-GPTQ-Int8 and LLaMA-3.1-8B-GPTQ-Q8 both stay near
        full-precision accuracy on Bangla NLU at roughly half the
        memory footprint, and both run comfortably on consumer GPUs
        (8–16 GB VRAM) without a meaningful accuracy cost.

  \item \textbf{Do not use GGUF-W8A16 where reasoning matters.} The
        GPT-OSS GGUF checkpoint we tested collapses on multi-step
        reasoning tasks. Until GGUF calibration for Bangla improves,
        we would not put this format in front of high-stakes
        reasoning tasks.

  \item \textbf{For comprehension-only use cases, quantization format
        matters less.} BoolQ-BN accuracy holds up across all three
        families, so applications built around passage-based fact
        retrieval can reasonably use any of the variants we tested.
\end{itemize}

\section{Conclusion}
\label{sec:conclusion}

We set out to evaluate post-training quantization on Bangla natural
language understanding across three LLM families, five benchmarks,
and several quantization formats. The main finding is that
quantization impact tracks architecture and format more than it
tracks language in general: GPTQ-Int8 and GPTQ-Q8 compression of
Qwen-2.5-7B and LLaMA-3.1-8B costs at most 1.5\% absolute accuracy
across all five benchmarks, which makes both configurations usable
for resource-constrained Bangla NLP deployment. GGUF-W8A16
quantization of GPT-OSS-20B is a different story: it degrades
reasoning and commonsense benchmarks by as much as 57.4\%, though
reading comprehension accuracy holds up reasonably well even there.

Reasoning tasks (CommonsenseQA-BN, OpenBookQA-BN) are more sensitive
to quantization than reading comprehension (BoolQ-BN) in all three
families we tested. This matches what prior work has found on
English benchmarks; we extend it, for what we believe is the first
time, to a low-resource, morphologically complex language.

\paragraph{Future work.}
A few directions stand out. INT4 formats (AWQ, Q4\_K\_M) would map
out the accuracy-compression trade-off for Bangla more fully than the
INT8-range formats we test here. Quantizing Bangla-native models such
as TituLLMs~\citep{nahin2025titullms} would show whether
language-specific pretraining adds any compression resilience on top
of what we see. Quantization-aware training for Bangla is worth
trying as a way to get more robust compressed models directly.
Inference latency and memory footprint benchmarking would round out
the accuracy picture we give here. And extending this work to
generative Bangla tasks, summarization, translation, dialogue, would
tell us whether the reasoning-vs-comprehension split we observe holds
outside classification-style benchmarks.


\section*{Ethics Statement}
All models and datasets used here are publicly available under
open-source licenses, and no private or personally identifiable data
is involved. The evaluations were run for academic research purposes
only. We do not expect direct harm from this work, but we do want to
flag one risk: a quantized model with degraded reasoning accuracy
could cause real problems if deployed in a high-stakes
decision-support setting without the kind of validation we describe
here.

\section*{Limitations}
We evaluate only INT8-range quantization formats; INT4 compression
(Q4\_K\_M, AWQ-4bit) may behave differently and is not covered here.
Zero-shot evaluation may not reflect performance in few-shot or
fine-tuned deployment scenarios. We do not measure inference latency
or peak memory usage, both of which matter for real deployment
decisions alongside accuracy. We were not able to include a quantized
Bangla-native model such as TituLLMs, so we cannot say whether
language-specific pretraining adds any extra quantization resilience
on top of what we observe here. Our benchmarks are all
classification-style; generative Bangla tasks such as summarization,
translation, or dialogue may show different sensitivity patterns
entirely.

There is also a design limitation worth stating plainly: family,
quantization format, and model size are not fully crossed in our
setup. GPT-OSS is the only family tested with GGUF, the only one at
20B parameters, and the only one without a GPTQ counterpart. This
means our central finding, that quantization impact tracks
architecture rather than language, cannot be fully separated from the
possibility that GGUF-W8A16 itself is simply less accurate than GPTQ
at this bit width, or that larger models calibrate worse. A full
$3\times3$ grid crossing family, format, and size would be needed to
pull these apart, and we leave that for future work. Finally, the
calibration dataset used for the GPT-OSS GGUF checkpoint is not
publicly documented, which adds a further confound to any
cross-family comparison involving that model.

\bibliography{references}

\appendix
\section{Dataset Statistics and Compute Budget}
\label{app:stats}

\Cref{tab:datastats} reports the number of examples in the evaluation
split used for each benchmark. All evaluation is zero-shot, so only
the test (or full) split is used.

\begin{table}[H]
\centering
\caption{Number of examples in the evaluation split for each
         benchmark.}
\label{tab:datastats}
\renewcommand{\arraystretch}{1.2}
\resizebox{\columnwidth}{!}{%
\begin{tabular}{lcc}
\toprule
\textbf{Benchmark} & \textbf{Task Category} & \textbf{\# Examples} \\
\midrule
Bangla MMLU       & Reasoning     & 729 \\
CommonsenseQA-BN  & Commonsense   & 1220 \\
OpenBookQA-BN     & Reasoning     & 497 \\
PIQA-BN           & Commonsense   & 1840 \\
BoolQ-BN          & Comprehension & 729 \\
\bottomrule
\end{tabular}%
}
\end{table}

Total compute for the study was approximately 40 GPU-hours across
all six model--precision configurations (three families $\times$
full-precision and quantized variants), evaluated over the five
benchmarks above; GPT-OSS-20B accounted for the largest share of this
budget given its parameter count.

\end{document}